\documentclass[letterpaper]{article} % DO NOT CHANGE THIS
\usepackage{aaai24}  % DO NOT CHANGE THIS
\usepackage{times}  % DO NOT CHANGE THIS
\usepackage{helvet}  % DO NOT CHANGE THIS
\usepackage{courier}  % DO NOT CHANGE THIS
\usepackage[hyphens]{url}  % DO NOT CHANGE THIS
\usepackage{graphicx} % DO NOT CHANGE THIS
\usepackage{natbib}  % DO NOT CHANGE THIS AND DO NOT ADD ANY OPTIONS TO IT
\usepackage{caption} % DO NOT CHANGE THIS AND DO NOT ADD ANY OPTIONS TO IT
\usepackage{amssymb}
\usepackage{multirow}
\usepackage{booktabs}
\usepackage{algorithm}
\usepackage{algorithmic}
\usepackage{easyReview}
\newcommand{\ie}{\textit{i}.\textit{e}., }

\usepackage{newfloat}
\usepackage{listings}
\usepackage{amsmath}
\DeclareCaptionStyle{ruled}{labelfont=normalfont,labelsep=colon,strut=off} % DO NOT CHANGE THIS
\floatstyle{ruled}
\newfloat{listing}{tb}{lst}{}
\floatname{listing}{Listing}
\title{Semantic Scene Graph Generation Based on an Edge Dual Scene Graph and Message Passing Neural Network}
\author{
Hyeongjin Kim\textsuperscript{\rm 1} \equalcontrib,
Sangwon Kim\textsuperscript{\rm 1} \equalcontrib,
Jong Taek Lee\textsuperscript{\rm 2},
Byoung Chul Ko\textsuperscript{\rm 1}\thanks{Corresponding author.}\\
}
\affiliations{
\textsuperscript{\rm 1}Dept. of Computer Engineering, Keimyung University\\
\textsuperscript{\rm 2}School of Computer Science and Engineering, Kyungpook National University\\
\{henryjoshuakim, eddiesangwonkim\}@gmail.com, jongtaeklee@knu.ac.kr, niceko@kmu.ac.kr
}

\usepackage{bibentry}
\nocopyright
\begin{document}

\maketitle

\begin{abstract}
	Along with generative AI, interest in scene graph generation (SGG), which comprehensively captures the relationships and interactions between objects in an image and creates a structured graph-based representation, has significantly increased in recent years. However, relying on object-centric and dichotomous relationships, existing SGG methods have a limited ability to accurately predict detailed relationships. To solve these problems, a new approach to the modeling multi-object relationships, called edge dual scene graph generation (EdgeSGG), is proposed herein. EdgeSGG is based on a edge dual scene graph and Dual Message Passing Neural Network (DualMPNN), which can capture rich contextual interactions between unconstrained objects. To facilitate the learning of edge dual scene graphs with a symmetric graph structure, the proposed DualMPNN learns both object- and relation-centric features for more accurately predicting relation-aware contexts and allows fine-grained relational updates between objects. A comparative experiment with state-of-the-art (SoTA) methods was conducted using two public datasets for SGG operations and six metrics for three subtasks. Compared with SoTA approaches, the proposed model exhibited substantial performance improvements across all SGG subtasks. Furthermore, experiment on long-tail distributions revealed that incorporating the relationships between objects effectively mitigates existing long-tail problems.
\end{abstract}

\begin{figure}[t]
	\centering
	\includegraphics[width=0.8\columnwidth]{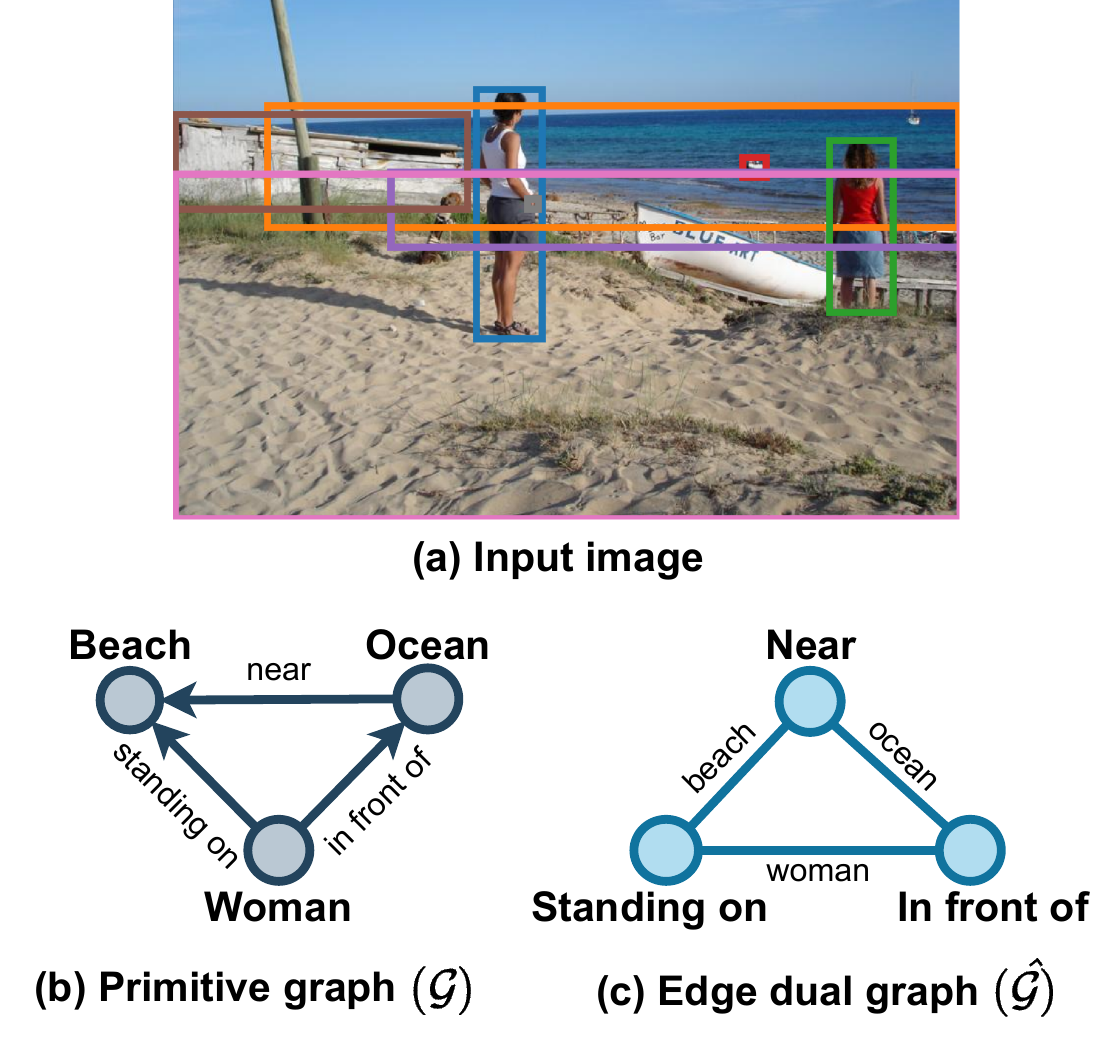}
	\caption{Process of generating the proposed edge dual scene graph. (a) Result of object detector from the input image. (b) A scene graph is formed using the objects as nodes and their relationship as edges extracted through (a). (c) The edge dual scene graph is transformed from the scene graph (b). The key idea of edge dual scene graph generation is to convert nodes into edges and edges into nodes.}
	\label{fig1}
\end{figure}

\section{Introduction}
Recent advancements in deep neural networks (DNNs) have resulted in an unprecedented performance in visual recognition tasks \cite{vqa, imgcap, star}. There has consequently been a continuous flow of research aimed at illustrating how DNNs aggregate and utilize visual information. An active area of research is scene graph generation (SGG), which uses DNNs to automatically map images onto semantically structured scene graphs.
To comprehensively capture the relationships and interactions between objects within an image, SGG primarily focuses on generating structured representations, frequently in the form of graphs. Therefore, SGG requires the correct labeling of detected objects and their relationships.
SGG approaches generally involve two primary steps: object detection and relationship modeling. Object detection is responsible for identifying and localizing individual objects within an image, resulting in a collection of bounding boxes. 
The subsequent step, relationship modeling, aims to comprehend the interactions among these objects and capture their contextual dependencies.
Several methods \cite{vctree, motifs} have been proposed to address the challenges to SGG, encompassing both rule- and learning-based approaches. Rule-based methods commonly employ predefined templates or heuristics to establish relationships between objects. Conversely, learning-based methods use DNNs to learn relationships from extensively annotated datasets. Graph neural networks (GNNs) are often employed to model structured representations and leverage graph-based operations for reasoning and prediction within an SGG framework.

Common design approaches \cite{bgnn, imp, hetsgg} in relationship modeling using GNNs aim to uncover the dependencies between objects by predicting graph edges in an object-centric manner. However, despite the need to construct a scene graph that considers the complex dependencies among various objects in a scene, most approaches in this field have focused on exploring the relationship between pairs of individual objects. For example, consider a scenario in which \textit{a person is riding a bike in a park}. Existing methods typically concentrate on dichotomous pairs such as  \verb|person|-\verb|bike|, \verb|bike|-\verb|park|, and \verb|person|-\verb|park|. Although an object in the real-world will often interact with multiple other objects, research to date has overlooked this aspect. Therefore, rather than solely considering simple object pairs, analyzing the mutual relationships among multiple objects can lead to an improved performance. Several methods \cite{general, relationformer, squat} have attempted to effectively model object interactions by analyzing the relationships among all graph edges. However, such exhaustive approaches incur an excessive number of computations owing to their comprehensive nature. HetSGG \cite{hetsgg} attempted to address the relationships among objects by proposing heterogeneous graphs for capturing the relationships in more detail. Nevertheless, the precise prediction of detailed relationships is limited because they rely on an object-centric approach.
To overcome the problems of dichotomous relationships and object-centric approaches, we propose edge dual scene graph generation (EdgeSGG), a novel relation-centric approach to modeling multiple object relationships for dependency prediction. As shown in Fig. {\ref{fig1}}, our method is based on the concept of an edge dual scene graph that allows the capture of rich contextual interactions between unconstrained objects. We demonstrate that EdgeSGG (i) facilitates fine-grained scene reasoning even in scenarios with complex objects, (ii) exhibits higher accuracy in predicting relationships than existing methods, and (iii) alleviates the long-tail problem. Finally, we illustrate how EdgeSGG can be utilized to enhance the understanding of DNNs by uncovering relationships using a reconstructed edge dual scene graph.

\section{Related Work}
\textbf{Scene Graph Generation.} Visual relationship \cite{lu2016} pioneered the more challenging task of detecting visual relations in the wild by modeling relationships in an independent manner. They adapted a separate relationship prediction scheme ({\ie} training only the relationship modeling function) to account for the infrequent nature of most relationships. 
\\
\textbf{SGG based on Contextual Information.} Many studies \cite{squat, penet, hetsgg, mdsn, motifs, imp} have been subsequently introduced to address the issue of ambiguity and to predict missing relationships by leveraging contextual information. Such studies are aimed at improving the accuracy of SGG in terms of relationship prediction by considering the interactions between objects in an image. IMP \cite{imp} first incorporated contextual information to enhance the relation modeling by utilizing an iterative message passing structure, which refining the object and relation features. A few recent studies have continued to explore prototyping \cite{penet} or transformer-based methods \cite{squat}, while also introducing approaches \cite{hetsgg} to generating diverse graphs for the synthesizing of rich contextual information from heterogeneous sources. However, most studies based on contextual information have continued to adopt object-centric approaches and have failed to adequately consider the unbalanced relationship distribution.
\\
\textbf{SGG based on Long-tail Solving.} Knowledg-embedded routing network \cite{chen2019} aimed to tackle the unbalanced distribution issue in wild scenes by incorporating statistical correlations between object pairs and their relationships. Following the study by \cite{chen2019}, several approaches \cite{nice, bgnn, tde} have attempted to solve the problem of an imbalanced class distribution. Although most of these studies have improved the SGG prediction performance of the tail classes in a long-tail data distribution (LTD), they have failed to significantly improve the overall performance, including the head.
To the best of our knowledge, the present study represents the first attempt to reconstruct a scene graph as an edge dual scene graph that effectively captures contextual information and employ a relation-centric approach to addressing the LTD problem.

\begin{figure*}[t]
	\centering
	\includegraphics[width=0.8\textwidth]{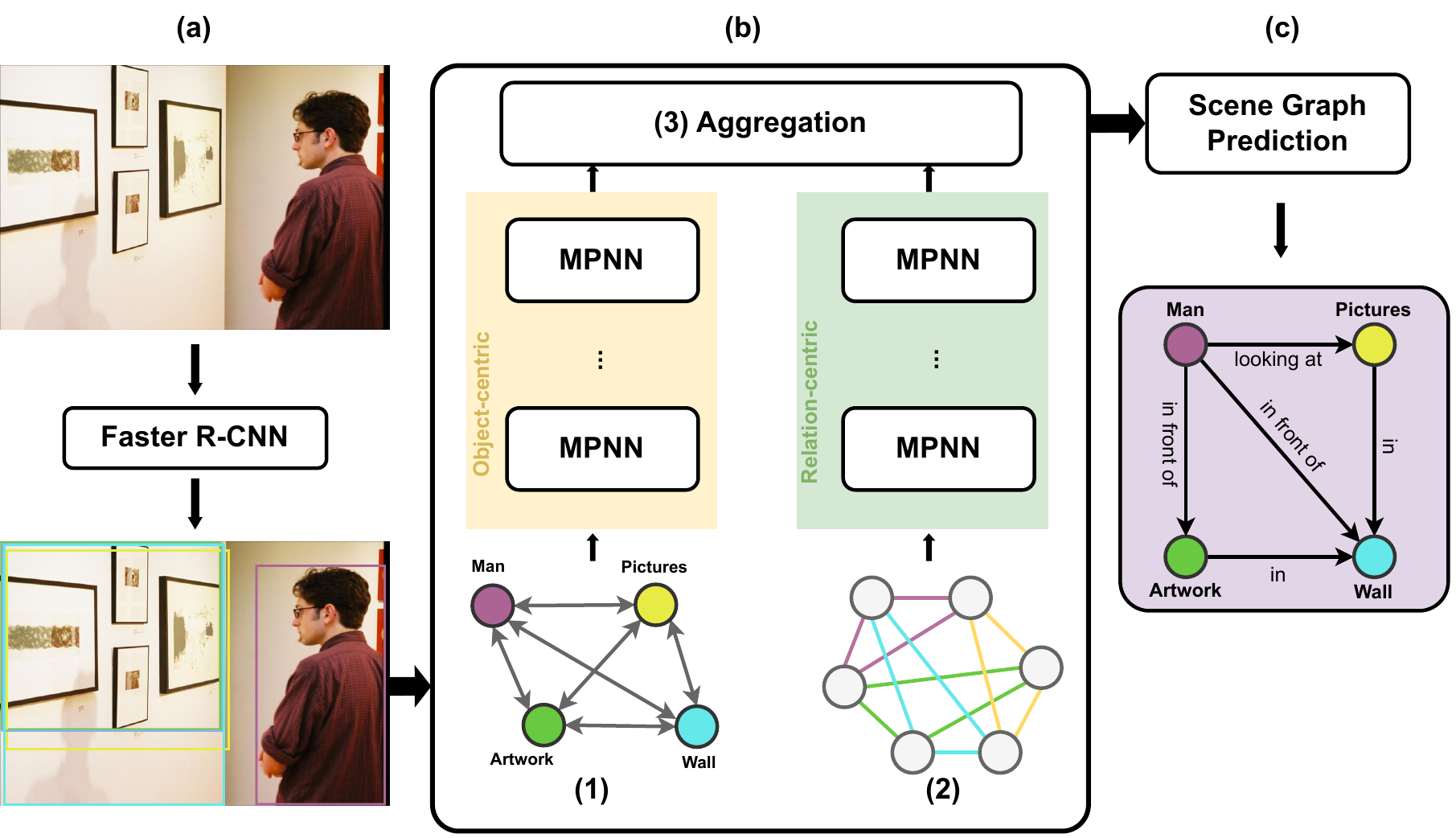} % Reduce the figure size so that it is slightly narrower than the column.
	\caption{(a) The object detection results are fed into DualMPNN. (b) The proposed DualMPNN consists of three parts: (1) object-centric MPNN, (2) relation-centric MPNN, and (3) feature aggregation. A primitive graph is formed to improve the object-centric learning from the results of the object detector, and at the same time, a symmetric edge dual scene graph is generated to enable improved relation-centric learning; the output of two MPNNs are then aggregated and fed to (c) the scene graph prediction module.}
	\label{fig2}
\end{figure*}

\section{Method}

\subsection{Preliminaries}
\noindent
\textbf{SGG Learning.} To be specific, within the independence paradigm, SGG learns a mapping from samples $x\in X $ to parsed scene graph $(\mathcal{O}, \mathcal{R}) \in \mathcal{T} $ by means of the following: (i) An object detection function $d:X \to (\mathcal{O},\mathcal{B}, L) \in \mathbb{R}^k$ maps a sample from the input space $x \in X \subseteq \mathbb{R}^n $ to the intermediate spaces $o\in \mathcal{O} \subseteq \mathbb{R}^{d_{o}}$, $b\in \mathcal{B} \subseteq {[0,1]}^4,$ and $l\in L \subseteq \mathbb{R}^{|Y|} $ formed by $k$ objects and their bounding boxes and classes. A symbol $d_{o}$ means the object feature dimension. (ii) A primitive graph building function $\psi :(\mathcal{O},\mathcal{B})\to \mathcal{G}$ maps samples from the object space $(o,b)\in (\mathcal{O},\mathcal{B})$ to a primitive graph $(\mathcal{N},\mathcal{E})\in \mathcal{G}$. Finally, (iii) a relationship modeling function $f:(\mathcal{N},\mathcal{E})\to \mathcal{R}$ maps samples from the object space $(o,l)\in (\mathcal{O},\mathcal{L})$ to a relationship space $r \in \mathcal{R} \subseteq \mathbb{R}^{d_r}$. Here, $d_r$ indicates relation feature dimension. During training, the SGG is encouraged to align $\hat{r} = f(d(x))$ with the predicted $(o,y)=d(x)$ to the corresponding ground-truth relation $r$ of $x$. This can be achieved by (i) employing the widely used ResNeXt-101-FPN \cite{resxt}, Faster R-CNN detector \cite{faster} and (ii) using its output to train the relationship-modeling function.

\noindent
\textbf{Dual Graph.} In graph theory, the dual graph of a planar graph $(\mathcal{N},\mathcal{E})\in \mathcal{G}$ has each face of $\mathcal{G}$ represented as a node. More precisely, given a graph $\mathcal{G}$ in which the edges do not overlap, it is possible to generate a dual graph that exhibits symmetry to {$\mathcal{G}$} by utilizing a mapping function $p:\mathcal{G}\to \hat{\mathcal{G}}$. However, in the case of a parsed scene graph  $\mathcal{T}$, where the edges may overlap depending on the interacting objects, it is impossible to directly generate a traditional dual graph. Therefore, inspired by \cite{edg}, we propose a novel approach, an edge dual scene graph. The original edge dual graph is a symmetric graph that preserves the primal graph structure while inverting the roles of each node and edge. By expanding original approach to image, it becomes possible to conduct relation-centric learning on different scenes, allowing for the learning of various object interactions and contextual information.

\subsection{EdgeSGG Overview}
We present an overview of our proposed framework for scene graph generation based on an edge dual scene graph with the dual message passing neural network (DualMPNN). As shown in Fig. {\ref{fig2}}, the proposed EdgeSGG comprises (a) an object detector, (b) a DualMPNN to enable object- and relation-centric learning with an edge dual scene graph generator, and (c) a scene graph prediction module.
\\
\noindent
\textbf{Object Detection Function.} Within a framework, the object detector (Fig. {\ref{fig2}} (a)) identifies the objects of interest in a scene. Following previous studies, we employ Faster R-CNN \cite{faster} as the object detector and Glove \cite{glove} as word embedding. As a result, the input image $x\in X$ is mapped to $(o,b,l)\in (\mathcal{O}, \mathcal{B}, L)$. We then construct an initial primitive graph $(\mathcal{U}, \mathcal{E}, \mathcal{V})\in \mathcal{G}$ from the detected objects with function $\psi:(\mathcal{O}, \mathcal{B})\to \mathcal{G}$. Primitive graph $\mathcal{G}$ is defined by three components: $u\in \mathcal{U} \subset \mathbb{R}^{d_{o}}$ representing the subjects, $v \in \mathcal{V} \subset \mathbb{R}^{d_{o}}$ representing the objects, and $e_{<u,v>} \in \mathcal{E}$ representing the relationships from subject to object. The feature vector of the relationship between $u$ and $v$ is extracted from the bounding box positions of $u$ and $v$, and their corresponding union boxes. Here, $\mathcal{U}$ and $\mathcal{V}$ are sets of object pairs derived from $\mathcal{O}$, and $\mathcal{V} \subseteq \mathcal{U}$ which $\mathcal{V}$ is a subset of $\mathcal{U}$ that does not contain itself.
\\
\noindent
\textbf{Building Edge Dual Scene Graph.} As previously mentioned, a conventional dual graph is useful for synthesizing rich contextual information; however, its applicability is limited to only planar graphs with non-overlapping edges. In particular, the primitive graph $\mathcal{G}$ contains complex relationships among multiple objects within a scene, making it a challenge to guarantee it to be a planar graph. In this study, we propose a new edge dual scene graph to facilitating the concept-oriented understanding of a dynamic scene structure. Edge dual scene graph allows us to capture the complex relationships among multiple objects within a scene, thereby ensuring the incorporation of abundant contextual information.
Our proposed edge dual scene graph forms dual nodes from existing edges and edge duals from the existing nodes. The edge dual scene graph  $(\hat{\mathcal{U}_{i}}, \hat{\mathcal{E}}, \hat{\mathcal{U}_{j}})\in \hat{\mathcal{G}}$ is defined as follows:

\begin{equation}
	\hat{\mathcal{U}} = \{e_i | e_i \in \mathcal{E}, i = 1,2...|\mathcal{E}|\}
	\label{eq1}
\end{equation}
\begin{equation}
	\hat{\mathcal{E}} = \{u_i = (e_i, e_j) | e_i \cap e_j = u_i \in \mathcal{U}, i \neq j, u_i \neq \o \}
	\label{eq2}
\end{equation}
\\
\indent
To facilitate the calculation of the edge ordered pair $u_i$, the properties $(e_i,e_j )=\{\{e_i \},\{e_i,e_j\}\}$ are converted into a collective structure and utilized. The edge dual scene graph $\hat{\mathcal{G}}$ produced using Eqs. {\ref{eq1}} and {\ref{eq2}} consistently transforms the existing nodes of graph $\mathcal{G}$ into edges, and the edges into nodes. An edge dual scene graph transformation is presented as follows: Using the objects and relationships extracted from the object detector, as shown in Fig. {\ref{fig2}} (b), we construct a complete graph comprising $|N|$ nodes and $|E| = |N|(|N| - 1)/2$ edges. The corresponding graph is transformed into an edge dual scene graph $\hat{\mathcal{G}}$ consisting of $|E|$ nodes and $|E_{dual}| = |E|(|N| - 2)$ edges. Our example graph (Fig. {\ref{fig2}} (b)) has 4 nodes and 6 edges. In the edge dual scene graph, we obtain 6 nodes and 12 edges. In this transformation, the originally adjacent edges become nodes in a dual graph and the two nodes are connected by an edge. This transformation ensures that the subgraphs of the connected nodes in the dual graph correspond to those in the original functional network. Furthermore, because $\hat{\mathcal{G}}$ reflects the relationships among relationships, it facilitates message passing. A detailed description of the MPNN method using graphs $\mathcal{G}$ and $\hat{\mathcal{G}}$ is presented in the following section.

\noindent
\textbf{Dual Message Passing Neural Network.} In this study, we generated an edge dual scene graph that is symmetric to the primitive graph, enabling improved relation-centric learning. To facilitate the learning of two symmetric graphs, we propose the novel {DualMPNN}. The proposed DualMPNN differs from existing message passing as it learns both object-centric \cite{bgnn} and relation-centric features, enabling a more accurate prediction of relation-aware contexts and fine-grained relationship updates between objects. A {DualMPNN} comprises the following three parts: (1) an object-centric MPNN, (2) a relation-centric MPNN, and (3) feature aggregation.

\noindent
\textbf{\textit{Part 1)} Object-centric MPNN.} To update the object-centric features using the primitive graph $\mathcal{G}$, which has nodes $\mathcal{U}$ and $\mathcal{V}$, and edge $\mathcal{E}$, the feature update process applied is as follows:
\begin{multline}
	e_{<u,v>}^{h+1} = e_{<u,v>}^{h} \ + \ \sigma(\alpha(u, v)e^h_{<u, v>}W_u \ + \\ \ (1-\alpha(u, v))e^h_{<v, u>}W_v)	
	\label{eq3}
\end{multline}

\noindent
where $e_{<u,v>}^{h+1} \in \mathbb{R}^{d_r}$ is the $(h+1)$\textit{th} $\in \mathbb{N}_H$ relation feature ({\ie} $H$  is the total number of layers), and $W_u$ and $W_v \in \mathbb{R}^{d_r \times d_r}$ are the weight matrices. The activation function $\sigma$ is the ReLU, and $\alpha(u,v)$ is an attention score operation computed using the weight matrix $W_{att} \in \mathbb{R}^{d_o}$ as follows:
\begin{equation}
	\alpha(u, v) = {{\mathrm{exp}(W^\intercal_{att}u)} \over {\mathrm{exp}(W^\intercal_{att}u) + \mathrm{exp}(W^\intercal_{att}v)}}
	\label{eq4}
\end{equation}

\noindent
\textbf{\textit{Part 2)} Relation-centric MPNN.} To update the relation features, an object-centric MPNN considers only the dependencies between objects $\mathcal{U}$ and $\mathcal{V}$, which limits its ability to capture the contextual information of the neighborhood. To address this limitation, we introduce a relation-centric MPNN that enables an update of the edge features by considering such relative neighborhood relationships. Using the edge dual graph $\hat{\mathcal{G}}$ constructed through Eqs. \ref{eq1} and \ref{eq2}, the relation-centric MPNN is applied as follows:

\begin{equation}
	z^0_{<e_i,e_j>} = u_iW_{o2e},\ \mathrm{with} \quad W_{o2e} \in \mathbb{R}^{d_o \times d_r} 
	\label{eq10}
\end{equation}
\begin{multline}
	z^{h+1}_{<e_i,e_j>} = z^{h}_{<e_i,e_j>} \ +\ \sigma(\sum_{e_j \in \mathcal{N}(e_i)}\alpha(e_i, e_j)z^{h}_{<e_i,e_j>}W_i \ + \\ \alpha(e_j, e_i)z^{h}_{<e_j,e_i>}W_j) 
	\label{eq5}
\end{multline}

\noindent
where $z_{<e_i,e_j>}^{h+1} \in \mathbb{R}^{d_r}$ is a relation-centric edge feature of the $(h+1)$\textit{th} MPNN layer, $W_i$, $W_j \in \mathbb{R}^{d_r \times d_r}$ and $W_{o2e}$ are the weight metrics. The relation-centric MPNN can incorporate not only simple $(u, v)$ pairs but also edge features with contextual neighborhood information, facilitating fine-grained scene graph generation. The effectiveness of this method is demonstrated through the long-tail solution shown in Fig. {\ref{fig3}}.

\noindent
\textbf{\textit{Part 3)} Feature Aggregation.} The relation features $e^H$ and $z^H$, generated using the two MPNN methods were combined via a concatenation operation. Subsequently, the concatenated feature is fed into a fully connected layer to derive the final feature vector $p_r$, which encompasses both object-centric and relation-centric features.
\begin{equation}
	p_r = \sigma(FC([e^H \parallel z^H]))
	\label{eq6}
\end{equation}

\noindent
Here, $\parallel$ indicates the concatenation operation, and $FC$ denotes a linear layer defined as $FC:\mathbb{R}^{(d_r+d_r)} \to \mathbb{R}^{d_r}$. For simplicity, normalization and bias are omitted from Eq. {\ref{eq6}}.

\noindent
\textbf{Scene Graph Prediction and Training.} The prediction of the relationship label is inferred through the object feature $u$ extracted from a Faster R-CNN and a simple linear classifier, {\ie} the feature vector $p_r$ passing through the DualMPNN module. The object features $u$ and $p_r$ were applied to the linear layer using {$softmax$}.
\begin{equation}
	\hat{u} = softmax(uW_{obi}), \quad \hat{p_r} = softmax(p_rW_{rel})
	\label{eq7}
\end{equation}

\noindent
Here, $W_{obj} \in \mathbb{R}^{d_o \times |Y_{obj}|}$ and $W_{rel} \in \mathbb{R}^{d_r \times|Y_{rel}|}$ are the weight matrix of the linear classifier, and to simplify the equation, the bias vector of the two layers is omitted. Finally,  $\hat{u}$ indicates $l \in L$ ({\ie} $\hat{u}$ is a feature for the object classification). $Y_{obj}$ and $Y_{rel}$ are object and relation labels set, respectively. The object loss $\mathcal{L}_{obj}$ and relation loss $\mathcal{L}_{rel}$ are learned to converge in the direction in which the joint loss $\mathcal{L}$ is minimized.
\begin{equation}
	\mathcal{L}_{obj} = {1 \over |\mathcal{V}|} \sum_{i=1}^{|\mathcal{V}|} \mathcal{L}_{CE}(y_i, \hat{u}_i), \quad
	\mathcal{L}_{rel} = {1 \over |\mathcal{E}|} \sum_{i=1}^{|\mathcal{E}|} \mathcal{L}_{CE}(s_i, \hat{p_r}_i)
	\label{eq8}
\end{equation}

\begin{equation}
	\mathcal{L} = \mathcal{L}_{obj}  + \mathcal{L}_{rel}
	\label{eq9}
\end{equation}

\noindent
Here, $y_i \in Y_{obj}$  and $s_i \in Y_{rel}$ indicate the ground-truth vectors of the object and relation labels, respectively. $\mathcal{L}_{CE}$ is the cross-entropy loss.

\begin{table*}[ht]
	\resizebox{\textwidth}{!}{
		\begin{tabular}{c|c|c|c|c|c|c}
			\toprule
			\multirow{2}{*}{\textbf{Method}} & \multicolumn{2}{c|}{\textbf{PredCls}} & \multicolumn{2}{c|}{\textbf{SGCls}} & \multicolumn{2}{c}{\textbf{SGGen}} \\ \cline{2-7} 
			& \textbf{mR@ 50 / 100} & \multicolumn{1}{l|}{\textbf{R@ 50 / 100}} & \textbf{mR@ 50 / 100} & \multicolumn{1}{c|}{\textbf{R@ 50 / 100}} & \textbf{mR@ 50 / 100}        & \textbf{R@ 50 / 100}        \\
			\hline \hline
			Motifis \cite{motifs} & 14.6 / 15.8          & 66.0 / 67.9         & 8.0 / 8.5          & 39.1 / 39.9          & 5.5 / 6.8          & 32.1 / 36.9         \\ 
			VCTree \cite{vctree}& 15.4 / 16.6           & 65.5 / 67.4         & 7.4 / 7.9           & 38.9 / 39.8         & 6.6 / 7.7          & 31.8 / 36.1         \\
			G-RCNN \cite{grcnn}& 16.4 / 17.2          & 65.4 / 67.2         & 9.0 / 9.5          & 37.0 / 38.5         & 5.8 / 6.6          & 29.7 / 32.8         \\
			MSDN \cite{mdsn}& 15.9 / 17.5    & 64.6 / 66.6     & 9.3 / 9.7         & 38.4 / 39.8          & 6.1 / 7.2         &  31.9 / 36.6         \\
			Unbiased \cite{unbaised}& 25.4 / 28.7        & 47.2 / 51.6         & 12.2 / 14.0          & 25.4 / 27.9          & 9.3 / 11.1          & 19.4 / 23.2         \\
			GPS-Net \cite{gpsnet}& 15.2 / 16.6          & 65.2 / 67.1         &  8.5 / 9.1         & 37.8 / 39.2         & 6.7 / 8.6          & 31.1 / 35.9         \\
			RU-Net \cite{runet}& - / 24.2          & 67.7 / 69.6         & - / 14.6          &  42.4 / 43.3        &  - / 10.8         & 32.9 / 37.5         \\
			R-CAGCN \cite{rcagcn}& 18.3 / 19.9          & 66.6 / 68.3         & 10.2 / 11.1          & 38.3 / 39.0         & 7.9 / 8.8          & 28.1 / 31.3         \\
			
			Nice-Motif \cite{nice}& 29.9 / 32.3        & 55.1 / 57.2         & 16.6 / 17.9          & 33.1 / 34.0          & 12.2 / 14.4          & 27.8 / 31.8         \\
			PPDL \cite{ppdl}& 32.2 / 33.3        & 47.2 / 47.6         & \underline{17.5} / 18.2          & 28.4 / 29.3          & 11.4 / 13.5          & 21.2 / 23.9         \\
			BGNN \cite{bgnn}& 30.4 / 32.9          & 59.2 / 61.3         &  14.3 / 16.5         &  37.4 / 38.5        & 10.7 / 12.6          & 31.0 / 35.8         \\
			IS-GGT \cite{isggt}& 26.4 / 31.9          &   - / -       & 15.8 / \textbf{18.9}          &  - / -        & 9.1 / 11.3          & - / -         \\
			HetSGG \cite{hetsgg} & 31.6 / 33.5          & 57.8 / 58.9         & 17.2 / 18.7          & 37.6 / 38.7         & 12.2 / 14.4          & 30.0 / 34.6         \\
			HetSGG++ \cite{hetsgg} & \underline{32.3} / \underline{34.5}          & 57.1 / 59.4         & 15.8 / 17.7          & 37.6 / 38.5         & 11.5 / 13.5         & 30.2 / 34.5         \\
			PENET \cite{penet}& 31.5 / 33.8          & 68.2 / 70.1       &  \textbf{17.8} / \textbf{18.9}       & 39.4 / 40.7       & \underline{12.4} / \underline{14.5}        &  30.7 / 35.2      \\ \toprule
			EdgeSGG (ours)& \textbf{34.7 / 36.9}        & 60.1 / 61.8       &  \textbf{17.8} / \underline{18.8}       &  39.1 / 40.1      & \textbf{13.6 / 15.8}          & 29.7 / 34.0         \\
			\bottomrule         
	\end{tabular}}
	\caption{Performance comparison with the SoTA SGG Methods on VG dataset. ++ denotes HetSGGplus \cite{hetsgg} model. \textbf{Bold} means best score, \underline{underline} means second-highest score}\label{table1}
\end{table*}

\begin{table}[ht]
	\resizebox{\columnwidth}{!}{%
		\begin{tabular}{c|c|c|c|c|c}
			\toprule
			\textbf{Method}	&\textbf{ mR@50} & \textbf{R@50} &  $\mathrm{\mathbf{wmAP}}_{rel}$ & $\mathrm{\mathbf{wmAP}}_{phr}$ & $\mathrm{\mathbf{score}}_{wtd}$ \\ \hline \hline
			RelDN \cite{reldn} & 37.2 & 75.3 & 32.2 & 33.4 & 42.0 \\
			VCTree \cite{vctree} & 33.9 & 74.1 & 34.2 & 33.1 & 40.2 \\
			G-RCNN \cite{grcnn} & 34.0 & 74.5 & 33.2 & 34.2 & 41.8 \\
			Motifs \cite{motifs} & 32.7 & 71.6 & 29.9 & 31.6 & 38.9 \\
			Unbiased \cite{unbaised} & 35.5 & 69.3 & 30.7 & 32.8 & 39.3 \\
			GPS-Net \cite{gpsnet}& 38.9 & 74.7 & 32.8 & 33.9 & 41.6 \\
			BGNN \cite{bgnn}& 40.5 & 75.0 & 33.5 & 34.1 & 42.1 \\
			HetSGG \cite{hetsgg}& 42.7 & 76.8 & 34.56& 35.5 & 43.3 \\
			HetSGG++ \cite{hetsgg}& 43.2 & 74.8 & 33.5 & 34.5 & 42.2 \\
			PENET \cite{penet}& - & 76.5 & \textbf{36.6} & \textbf{37.4} & \textbf{44.9} \\ \hline
			EdgeSGG (ours)& \textbf{43.3} & \textbf{77.1} & 36.4 & \textbf{37.4} & \textbf{44.9} \\ \hline
	\end{tabular}}
	\caption{Performance comparison with the SoTA methods on OI dataset.}	
	\label{table2}
	\vspace{-0.5cm}
\end{table}

\section{Experiments}

\noindent
\textbf{Datasets.} The proposed EdgeSGG enables relation-centric learning through edge dual graphs in public datasets for SGG tasks. Model training and evaluation were conducted using two types of datasets.

\begin{itemize}
	\item \textbf{Visual Genome (VG)} dataset \cite{vg} contains 108k images, with detailed annotations of objects and their relationships. Each image was annotated with bounding boxes and class labels for an average of 150 objects and 50 relationship labels for object pairs. The dataset presents a challenge owing to its long-tailed distribution of relationships. For a fair evaluation, the VG dataset is commonly split into training, validation, and testing sets.
	\item \textbf{OpenImages V6 (OI)} dataset \cite{oi6} is a large-scale dataset commonly used for SSG tasks. It contains a diverse collection of over 133k images with 126,368 training, 1,813 validation, and 5,322 testing images. This dataset covers a wide range of real-world scenarios. The OI provides object-level annotations for each image, including bounding boxes and 301 object categories. In addition, it includes 31 relationship annotations that describe the interactions and connections between pairs of objects within a scene.
\end{itemize}

\noindent
\textbf{Subtasks.} The SGG task can be divided into four main subtasks:

\begin{itemize}
	\item \textbf{Predicate Classification (PredCls)} focuses on classifying the relationships between two objects in each image. The input to this task was an image along with object proposals that included ground-truth bounding boxes and labels.
	\item \textbf{Scene Graph Classification (SGCls)} uses image and object proposals as inputs, similar to PredCls. However, in contrast to PredCls, object proposals contain only ground truth bounding boxes without labels.
	\item \textbf{Scene Graph Generation (SGGen)} is the most challenging subtask because it simulates real-world conditions. It requires the model to detect objects ({\ie} determine the bounding boxes and labels) in an image and predict the relationships between them.
\end{itemize}

\noindent
\textbf{Metrics.} In our experiments, we employed two key evaluation metrics ({\ie} recall@K and Mean Recall@K) to assess the performance of the proposed EdgeSGG and compare it with previous SoTA approaches. For all of the experiments, we calculated the metrics for K $ \in \{50,100\}$.

\begin{itemize}
	\item \textbf{Recall@K (R@K)} is a commonly used metric in SGG tasks that measures the ability of a model to predict at least one correct relationship among the top K predicted relationships for each object in a given scene.
	\item \textbf{Mean Recall@K (mR@K)} is the average of R@K scores across all objects in a scene. This provides a more comprehensive evaluation of the ability of the model to predict relationships for different objects in a scene.
\end{itemize}
In addition to these two metrics, we employed the following three additional metrics to provide a more comprehensive assessment of the SGG methods:

\begin{figure*}[ht]
	\centering
	\includegraphics[width=1\textwidth]{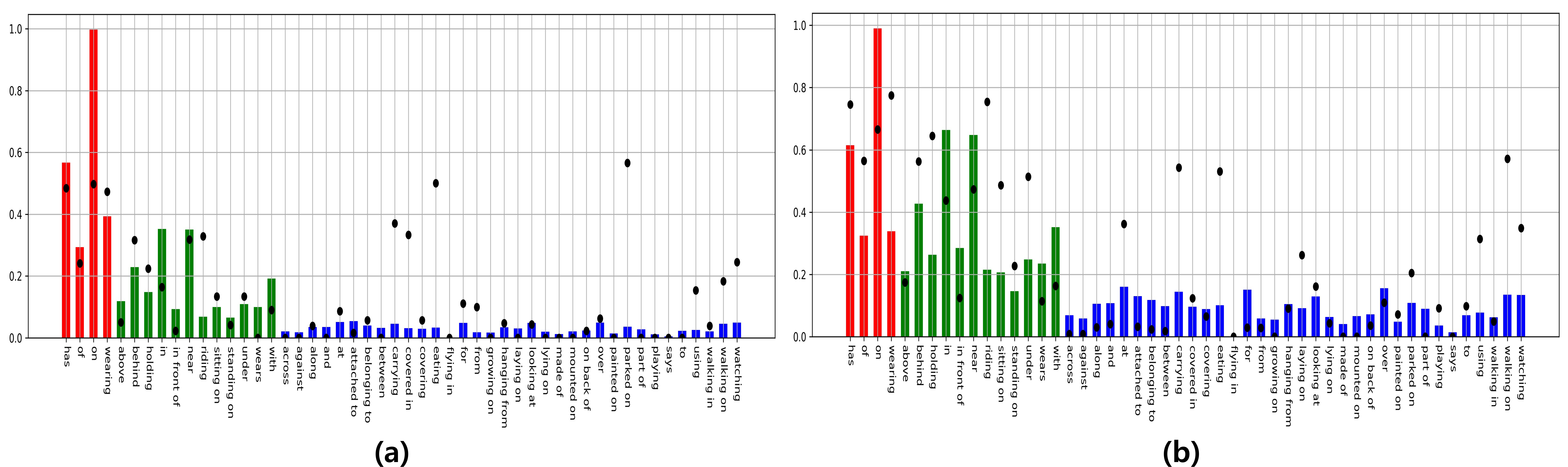} % Reduce the figure size so that it is slightly narrower than the column.
	\caption{Predicate prediction changes in VG dataset according to learning progress. (a) is the predicate result of iteration = 1500 and (b) is the predicate result of iteration = 49500. The \textbf{red} bar represents head, the \textbf{green} bar represents body, and the \textbf{blue} bar represents tail label. The \textbf{dot} represents the recall of each label.}
	\label{fig3}
	\vspace{-0.5cm}
\end{figure*}

\begin{itemize}
	\item \textbf{Weighted Mean Average Precision for Relationships ($\mathrm{\mathbf{wmAP}}_{rel}$)} evaluates the performance of the model in predicting the relationships between object pairs. It calculates the mean AP for each relationship category, weighted by the number of ground-truth instances of that relationship in a dataset. It provides a more balanced evaluation by considering the varying frequencies of different relationship types in scene graphs.
	
	\item \textbf{Weighted Mean Average Precison for Phrases ($\mathrm{\mathbf{wmAP}}_{phr}$)} assesses the ability of the model to predict relationship phrases involving both object categories and their corresponding relationships.
	
	\begin{table}[t]
		\begin{center}
			\begin{small}
				\begin{tabular}{cc|ccc}
					\toprule
					\multicolumn{2}{c|}{\textbf{Learning}}                          & \multicolumn{3}{c}{\textbf{SGGen}}                                                   \\
					\multicolumn{1}{c}{\textbf{Object}} & \multicolumn{1}{c|}{\textbf{Relation}} & \multicolumn{1}{c}{\textbf{mR@20}} & \multicolumn{1}{c}{\textbf{mR@50}} & \multicolumn{1}{c}{\textbf{mR@100}} \\ \hline \hline
					$\checkmark$  	&                      &    9.3                   &    12.2                   &    14.0                    \\ 
					&         $\checkmark$               &        9.9              &              12.7         &         14.7                \\ 
					$\checkmark$ 	&       $\checkmark$                  &        \textbf{ 10.4  }           &    \textbf{ 13.6   }               &     \textbf{  15.8  }     \\  \hline   
				\end{tabular}
			\end{small}
		\end{center}
		\caption{Performance impact of the proposed edge dual scene graph of DualMPNN.}
		\label{table3}
	\end{table}
	
	\item \textbf{Weighted Score ($\mathrm{\mathbf{score}}_{wtd}$)} is a comprehensive evaluation metric that combines the performance of the model with both the relationship and phrase predictions, considering their relative importance in scene graphs. This is the weighted sum of $\mathrm{\mathbf{wmAP}}_{rel}$ and $\mathrm{\mathbf{wmAP}}_{phr}$, where the weights are determined based on the significance of the relationships and phrases in a dataset. $\mathrm{\mathbf{score}}_{wtd}$ was calculated as: $\mathrm{\mathbf{score}}_{wtd}$ = 0.2 × R@50 + 0.4 × $\mathrm{\mathbf{wmAP}}_{rel}$ + 0.4 × $\mathrm{\mathbf{wmAP}}_{phr}$. 
\end{itemize}

\noindent
\textbf{Setup.} All experiments were conducted on a private machine equipped with two Intel(R) Xeon(R) CPUs, {\ie} Gold 6230R CPU @ 2.10GHz; 128GB RAM, and four NVIDIA RTX 3090 GPUs. We used the SGD optimizer, the detailed settings and hyperparameters of which can be found in \textbf{Appendix A}.

\subsection{Quantitative Experiments}
\noindent
\textbf{Performance comparison on VG dataset.} Table {\ref{table1}} shows the measured performances using four key metrics, mR@50/100 and R@50/100, for each subtask. Our proposed EdgeSGG demonstrated an outstanding performance across all subtasks and mR@K metrics when compared with the SoTA approaches. Because the VG dataset has an imbalanced data distribution, mR@K, which prefers tail predicates, can be said to be more reliable than R@K metrics that focus on common predictions with abundant samples. Although EdgeSGG has slightly lower scores for R@50 and 100 than other SoTA methods, its relatively high score for mR@K indicates that it is also robust to imbalanced data distributions.
For \textbf{PredCls}, EdgeSGG achieved 2.4\% higher mR@50/100 scores than the second-highest method, HetSGG, indicating its effectiveness and generic capture of more relevant predicates within the top-50 and top-100 predictions, respectively. Similarly, for \textbf{SGCls}, our EdgeSGG showed the highest score for PENET at mR@50/100. 
The ability of EdgeSGG to capture fine-grained scene graph classifications is evident from its superior performance among all metrics. In the case of \textbf{SGGen}, the proposed EdgeSGG outperformed all the other methods in terms of mR@50/100. The results indicate the robustness of our method in accurately and consistently detecting scene graph relationships. More importantly, when comparing \cite{gpsnet, bgnn, hetsgg} which are the same paradigms as the MPNN methods, we observed that the proposed EdgeSGG is superior among all subtasks. This indicates that for effective scene graph prediction, the consideration of not only object-centric MPNNs, but also relation-centric MPNNs, helps improve the performance.
The consistent superiority of our method in capturing fine-grained relationships and contextual information validates its potential in advancing SoTA approaches used in SGG on the VG dataset.

\begin{table}[t]
	\begin{center}
		\begin{small}
			\begin{tabular}{c|ccc}
				\toprule
				\multicolumn{1}{c|}{\multirow{2}{*}{\textbf{Variants}}} & \multicolumn{3}{c}{\textbf{SGGen}} \\
				\multicolumn{1}{c|}{}                   & \textbf{mR@20}     & \textbf{mR@50}     & \textbf{mR@100}    \\ \hline \hline
				Mean &  9.7     &  12.7     &  15.0     \\
				Multiple &  10.1     &  12.9     &  15.1     \\
				Concat &  \textbf{10.4}     &  \textbf{13.6}     &   \textbf{15.8}    \\ \hline
			\end{tabular}
		\end{small}
	\end{center}
	\caption{Performance changes according to the proposed feature aggregation function type.}
	\label{table4}
\end{table}

\begin{figure*}[ht]
	\centering
	\includegraphics[width=0.8\textwidth]{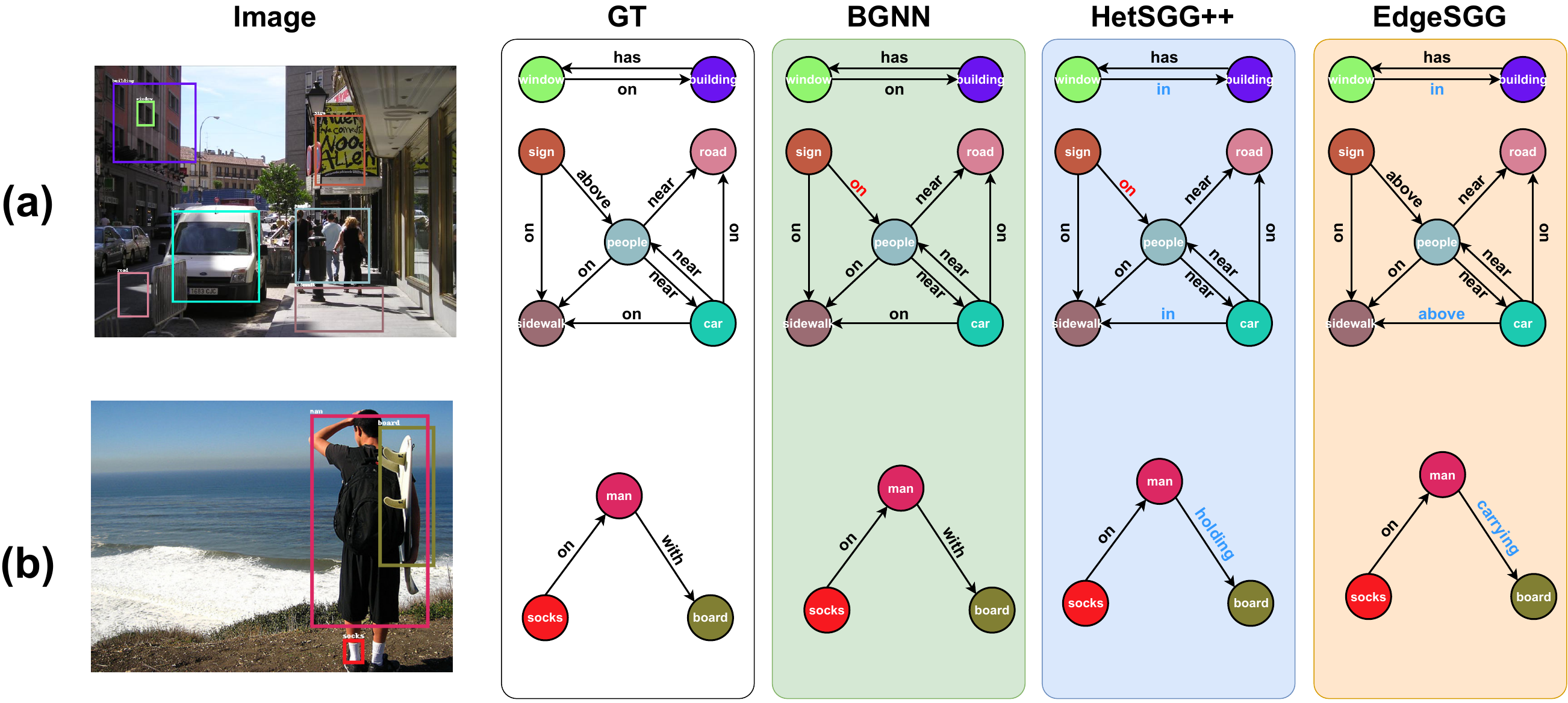} % Reduce the figure size so that it is slightly narrower than the column.
	\caption{Quantitative performance comparison for VG dataset. We confirm that the relation-centric MPNN of EdgeSGG enables a more fine-grained scene graph prediction than the comparison methods. The assessment was conducted on the PredCls subtask. (\textbf{Red}: incorrect predictions, \textbf{Blue}: correct predictions which more fine-grained)}
	\vspace{-0.3cm}
	\label{fig4}
\end{figure*}

\iffalse
\begin{figure*}[t]
	\centering
	\includegraphics[width=0.8\textwidth]{fig4} % Reduce the figure size so that it is slightly narrower than the column.
	\caption{Result of scene graph prediction of the proposed EdgeSGG on three subtasks using a dataset. (a) Examples of scene graph generation, (b) predicate classification, and (c) predicate classification.}
	\label{fig}
\end{figure*}
\fi 

\noindent
\textbf{Analysis on long-tail distribution.} In Fig. {\ref{fig3}}, we present a performance analysis of the proposed EdgeSGG on the long-tail distribution. Figure {\ref{fig3}} (a) shows a performance bar chart for each class during an early training iteration, and Fig. {\ref{fig3}} (b) displays a performance bar chart for a late training iteration. The bars in the charts represent the number of accurately classified objects for each class, providing insight into the performance of the model for different classes. In addition, the dotted charts placed above each bar indicate the recall for each class, representing the ability of the model to correctly predict relationships for objects within that class. Analyzing Fig. {\ref{fig3}} (a), we can observe that classes with fewer instances have lower bars, indicating that the model faces challenges in accurately predicting these less frequent classes during the early training phase. However, as the training progresses, Fig. {\ref{fig3}} (b) shows a significant improvement, with higher bars for the previously underrepresented classes. This improvement was not limited to the tail, but was also found in the head and body parts, confirming that the proposed EdgeSGG provides an overall performance improvement. Moreover, the dot charts above each bar in Fig. {\ref{fig3}} (b) show increasing recall scores across classes, further highlighting the ability of the model to generalize and predict relationships for objects in the long-tail classes.

\subsection{Ablation Studies}
\textbf{The effect of edge dual scene graph.} In this section, we describe an ablation study conducted to investigate the performance impact of the proposed edge dual scene graph in the SGDet subtask. Table {\ref{table3}} presents the evaluation results, with each row corresponding to two settings: object- and relation-centric learning of DualMPNN. Each column represents the evaluation metric mR@20/50/100 for the SGDet subtask. In object-centric learning, we train the entire architecture without utilizing the proposed edge dual scene graph in the DualMPNN, whereas relation-centric learning uses an edge dual scene graph without object-centric learning. Table 3 clearly shows that when we used object- and relation-centric learning together, the performance was consistently better than when we used only object- or relation-centric learning for all metrics. This result highlights the significant performance improvement achieved by integrating the object- and relation-centric aspects of the edge dual scene graph in DualMPNN. This ablation study demonstrated that the edge dual scene graph plays a vital role in boosting the performance of EdgeSGG. By leveraging relation-centric learning through the edge dual scene graph, the proposed EdgeSGG outperforms the object-centric MPNN.

\noindent
\textbf{Feature aggregation strategies.} To determine the optimal method for aggregating objects and relation-centric features, we conducted an ablation study using a feature combination method. Table {\ref{table4}} presents a performance comparison of three methods, {\ie}mean, multiplication, and cat, for the SGGen task. The mean and multiplication methods were 0.7\%, 0.9\% and 0.8\%, and 0.3\%, 0.7\% and 0.7\% lower for mR@20/50/100 in comparison to the Concat method, respectively. This can be attributed to the inherent nature of the mean and multiplication methods, which mix rather than preserve the features. By contrast, the Concat method demonstrated a superior performance because it was able to construct a feature vector that effectively reflects the characteristics of each individual feature.

\subsection{Visualization}
\iffalse
\textbf{Edge SGG Result.} To demonstrate the quantitative results of the proposed scene graph generation method, we visualized the prediction results for various subtasks. Figures \replace{4}{\ref{fig4}} (a) and \replace{4}{\ref{fig4}} (b) show the results of the scene graph generation task. In these two examples, even if only some objects were detected, the relationship between the objects did not differ significantly from the ground truth. The second row represents the results of the predicate (Fig. \replace{4}{\ref{fig4}} (b)) and scene graph (Fig. \replace{4}{\ref{fig4}} (c)) classification. Both subtasks predict the relationships between objects well, even if multiple objects are detected in the image. This reflects the relationship between objects and relationships through the proposed relation-centric MPNN.
\fi

\noindent
\textbf{Compare with SoTA method.} To obtain more convincing quantitative results, we compared the predicted scene graph results of EdgeSGG with those of BGNN, HetSGG, an MPNN paradigm for SoTA models. As illustrated in Fig. {\ref{fig4}}, the proposed method updates the feature by considering the relationships between relationships, making it possible to predict a finer-grained scene graph. In particular, in relation of the \verb|man|-\verb|board| in Fig \ref{fig4} (b), we can see that the predicate ``\textit{carrying}'' is more detailed than the previous methods.

\section{Conclusions}
We proposed EdgeSGG, an MPNN based on an edge dual scene graph, and a novel method for scene graph generation. We demonstrate that the proposed method outperforms the SoTA SGG models on the benchmark datasets. The proposed relation-centric MPNN method is applicable to various graph and SGG tasks. However, the computational cost of generating an additional edge dual scene and message passing process is considered a limitation of the proposed method. Therefore, in a future study, we plan to develop a more efficient edge dual scene graph that reduces the computational cost of message passing and can be applied to various scene interpretation tasks.

\bibliography{aaai24}

\end{document}